%% file: main.tex
\documentclass[10pt, a4paper, oneside, reqno]{amsart}

\input{style_P}
\input{command}

\usepackage{url}            

\usepackage{algorithm}
\usepackage{algorithmic}

 \usepackage[foot]{amsaddr}

\usepackage{subfig}

\usepackage{algorithmic}
{

{
		\theoremstyle{plain}
		
	}
	{
		\theoremstyle{plain}
		
	}
	{
	\theoremstyle{plain}
	
}
	{
		\theoremstyle{plain}
		
	}
	{
		\theoremstyle{plain}
		
	}
{
		\theoremstyle{plain}
		
	}
{
		\theoremstyle{plain}
		
	}}

\title[]{Multi-Agent Reinforcement Learning with Graph Convolutional Neural Networks for optimal Bidding Strategies of Generation Units in Electricity Markets }

\author[1]{Pegah Rokhforoz$^1$}
\author[2]{Olga Fink$^{2}$}
\address[1]{Chair of Intelligent Maintenance Systems, ETH Zurich, Switzerland.}
\address[2]{Laboratory of Intelligent Maintenance and Operation Systems, EPFL, Switzerland. Corresponding author, email address:olga.fink@epfl.ch.}

\begin{document}

\maketitle

\begin{abstract}
Finding optimal bidding strategies for generation units in electricity markets would result in higher profit. However, it is a challenging problem due to the system uncertainty which is due to the unknown other generation units' strategies. Distributed optimization, where each entity or agent decides on its bid individually, has become state of the art. However, it cannot overcome the challenges of system uncertainties. Deep reinforcement learning is a promising approach to learn the optimal strategy in uncertain environments. Nevertheless, it is not able to integrate the information on the spatial system topology in the learning process. This paper proposes a distributed learning algorithm based on deep reinforcement learning (DRL) combined with a graph convolutional neural network (GCN). In fact, the proposed framework helps the agents to update their decisions by getting feedback from the environment so that it can overcome the challenges of the uncertainties. In this proposed algorithm, the state and connection between nodes are the inputs of the GCN, which can make agents aware of the structure of the system. This information on the system topology helps the agents to improve their bidding strategies and increase the profit. We evaluate the proposed algorithm on the IEEE 30-bus system under different scenarios. Also, to investigate the generalization ability of the proposed approach, we test the trained model on IEEE 39-bus system. The results show that the proposed algorithm has more generalization abilities compare to the DRL and can result in higher profit when changing the topology of the system.
\end{abstract}


\section{Introduction}
Developing optimal bidding strategies for the strategic generation units plays a crucial role in increasing the electrical market profit. In this system, the generation units aim to find their optimal bidding strategies to participate in the electrical market. The traditional centralized decision-making with a large number of units encounters several challenges. First, the central system needs to have the full information of all the units. Therefore, the privacy of the units is not preserved. Second, the central system has to solve a large-scale optimization problem to maximize the reward, which requires high computational time. 

To address these challenges, distributed decision-making has been increasingly applied in this field. With distributed decision-making, each unit, referred to as an agent, obtains its decisions individually, and then shares them with other agents \cite{trentesaux2009distributed}. This problem typically is solved using non-cooperative game theory approach. Although, this approach has many advantages, such as preserving the private information and reducing the computational time, it still remains a very challenging problem. This is especially true when the complexity of the system increases, such as in the case of network systems \cite{yang2019survey,chen2021accommodating,rokhforoz2021multi}. Following challenges need to be solved for distributed decision-making in network systems: 1) In network systems, the agents decide non-cooperatively and their decision can influence the entire network. This requires the application of non-cooperative game theory. Nash Equilibrium (NE) can be considered as a feasible solution of this game \cite{nash1951non}. It describes a situation where no agent can increase its profit by changing its strategy as long as the other agent does not change its strategy \cite{song2002nash}. Nash equilibrium (NE) describes a situation that no agent can increase its profit by changing its strategy as long as the other agents are not changing their strategies. 2) When the system faces many uncertainties as is the case in electrical market bidding and the conditions or states of the system change dynamically, it may be difficult to find the NE of the game in a distributed manner \cite{sastry1994decentralized}.

In the electricity market, the generation units who participate in the market decide about their bidding decisions. In this system, the central system which is known as independent system operator (ISO) is responsible to clear the market, calculates the price of the market, and determines the power generation of each unit such that the load demand of the system is satisfied. In this system, each agent's reward depends on the market price set by an ISO. In fact, agents find their strategy which can maximize the reward in the market, and then ISO clears the market to minimize the cost of the system. Indeed, finding the optimal bidding strategies can be formulated as a bi-level optimization problem, where at the first level the agents decide their bidding, and at the second level, the ISO determines the price of the market after receiving the  agents' bid \cite{weber1999two, shafiekhani2019strategic}. In this market, since each agent's reward depends on the price of the market, which is the function of all other agents' decisions, we face a non-cooperative game. Furthermore, since the agents know neither the decisions of other agents nor the market price, there are certain system uncertainties \cite{li2005strategic}.

Several research studies have addressed the problem of distributed optimal bidding in the electricity market using the non-cooperative game approach \cite{ye2019incorporating, dai2016finding,wang2017strategic,zhang2010restructured,pozo2011finding}. However, previous research, system uncertainty has not been considered. Moreover, previous research did not propose a learning algorithm which converges to the NE of the game.

Reinforcement learning (RL) is a promising approach to solve the problem with uncertainty and incomplete information \cite{sutton2018reinforcement}. In RL, agents learn other agents' behavior and strategies by receiving environmental feedback. Q-learning is a model-free, off-policy reinforcement learning algorithm. Given the current state, it will find the best course of action. It typically implements look-up table to find agents' optimal strategy in each state of the system \cite{watkins1992q,jogunola2020consensus}. However, since the state and strategy space of the agents are continuous, this approach has a high computational time. To solve this challenge, the look-up table can be replaced by neural networks, in which two deep neural networks named actor and critic are used to estimate the agents' strategy and value function \cite{arulkumaran2017deep,xu2019deep}. Several studies addressed the problem of optimal bidding based on the actor-critic methodology \cite{du2021approximating,liang2020agent,ye2019deep,rokhforoz2021safe,rokhforoz2020agent,yin2020review}.

However, none of the previous works explicitly consider the network topology of the system when designing the deep RL (DRL) algorithm. One way to solve this challenge is to combine the graph convolutional neural networks (GCN) with the RL algorithm. GCN provides powerful topology embedded feature extraction capabilities \cite{kipf2016semi}.

In this paper, we propose a novel RL algorithm combined with GCN to obtain the optimal bidding strategy of the agents. The proposed approach is able to cope with the challenges of incomplete information. It is also computationally efficient and able to include the architecture of the network to find the optimal agents' policy. With the proposed approach, the generation units can find their optimal strategies without knowing the bidding strategies of other units. The proposed algorithm enables them to find their strategies by receiving feedback and rewards from the electricity system and ISO. We evaluate the proposed algorithm on IEEE 30-bus system with different capacities of the units. Moreover, we also evaluate the generalization ability of the proposed framework by training the model on data from IEEE 30-bus system and applying it on the IEEE 39-bus system, again with different unit capacities. The simulation results show that the proposed approach can achieve a higher gain compared to DRL with feed-forward neural networks and also has better generalization abilities. 

The main contributions of the paper are as follows:

1) We formulate the power bid strategies of generating units as a two-level optimization problem where the agents decide on their bidding decisions on the first level and then send them to ISO. At the second level, ISO clears the market and obtains the market price and power generation of each unit in a way that reduces the overall cost of the system.

2) We propose a multi-agent DRL algorithm to obtain the bidding strategies of the units. This approach can handle the challenges of uncertainties and the system's incomplete information. Moreover, it is computationally tractable since it applied a neural network rather than a look-up table to estimate the value function and the agents' strategies.

3) We propose to use GCN as the critic and actor networks to estimate the value function and policy of each agent. The proposed approach can consider the network architecture of the smart grid system and has a better generalization capability compare to DRL.

The rest of the report is organized as follows. The preliminaries of RL methodology and the proposed RL algorithm are introduced in Section \ref{sec:preliminary}. The units' and ISO's objective functions are formulated in Section \ref{sec:problem}. The solution based on a multi-agent RL combined with a GCN is proposed in Section \ref{sec:solution}. Case study evaluations are presented in Section \ref{sec:case study}. Concluding remarks are made in Section \ref{sec:conclusion}.

\subsection{Indices and sets}
$\mathcal{N}$: set of units ${\{1,\cdots,N}\}$ indexed by i. 

$\mathcal{N}_{s}$: set of samples ${\{1,\cdots,N_{s}}\}$ indexed by j.

$\mathcal{T}$: set of operational intervals ${\{1,\cdots,T}\}$ indexed by t.

\subsection{Parameters}
$t$: Number of time period 

$\powermax$: maximum power generation of generation unit $i$ ($MW$)

$\powermin$: minimum power generation of generation unit $i$ ($MW$)

$\marginalcost$: marginal cost of generation unit $i$ ($\frac{\$}{MW}$)

$\biddingmax$: maximum bidding of generation unit $i$

\subsection{Variables}

$\powera$: power generation of generation unit $i$ at time $t$ ($MW$)

$\price$: electricity market price at time $t$ ($\frac{\$}{MW}$)

$\biddingt$: bidding decision of generation unit $i$

\section{Preliminary}
\label{sec:preliminary}

\subsection{Deep Reinforcement learning}
In the following, we formulate the aim of DRL approach. Let us define the states and action space as $\mathcal{S}$ and $\mathcal{A}$, respectively. At time $t$ and given the state  $s(t)\in\mathcal{S}$, the agent chooses the action $a(t)\in\mathcal{A}$ which makes the transition to the new state $s(t+1)\in\mathcal{S}$ and provides a reward $r$. The objective function of each agent is to find the policy $\pi:\mathcal{S}\to\mathcal{A}$ that can maximize the expected reward. Let us define an action-value function as follows:

\begin{equation}
    Q_{\pi}(s({t}),a({t}))= {\mathbb E}_{\pi}\Big[\sum\limits_{k=0}^{N_{T}}\gamma^{k}r({t+k+1})|s({t}),a({t}),\pi\Big],
\end{equation}
where $\gamma\in[0,1)$ is the discount factor. The expected value of the reward over policy $\pi$ is given by ${\mathbb E}_{\pi}[\cdot]$. Hence, we can formulate the objective of RL agent as follows:

\begin{equation}
    Q^{*}(s_{t},a_{t})= \max\limits_{\pi}Q_{\pi}(s_{t},a_{t}).
    \label{eq:q learning}
\end{equation}
Q-learning is a model-free, off-policy reinforcement learning algorithm that can solve this problem by storing the action value function of all possible states and actions using a lookup table \cite{watkins1992q}. However, it faces the challenges of the "curse of dimensionality" when either the state or the action is continuous. Deep Q-Learning is one way to solve these challenges, using deep neural networks to estimate the value function and the corresponding policy.

\subsection{Graph convolutional neural network}
A graph consists of nodes and edges, which can be represented by $G=(V, E)$ and characterized by a feature matrix $X$ and a graph adjacency matrix $A$. Suppose the graph has $N$ vertices in which each vertex contains $f$ features denoted by $x_{i}$, $i=1,\cdots,N$. In this case, the feature matrix $X$ is a matrix with dimension $N\times{f}$, where the row $i$ contains the features of node $i$. The adjacency matrix is an $N\times{N}$ matrix that represents the graph structure. In this matrix, if node $i$ is connected to node $j$, then the element of row $i$ and column $j$ is one, otherwise, it is zero. The input of the GCN layer is the tuple ${\{X, A}\}$. For a GCN with $L$ layers, the output of layer $l$, $l=1,\cdots,L$, can be formulated as a non-linear function \cite{scarselli2008graph,welling2016semi}:

\begin{equation}
O^{l+1}=g(O^{l},A),
\end{equation}
where the input of the first level $O^{0}$ is the feature matrix $X$. The nonlinear function is denoted by $g$ and can be defined as follows:

\begin{equation}
g(O^{l},A)=\sigma({\hat{D}}^{-\frac{1}{2}}\hat{A}\hat{D}^{-\frac{1}{2}}O^{l}W^{l}),
\end{equation}
where $W^{l}$ corresponds to the weight matrix of layer $l$, $\sigma$ is the nonlinear function, $\hat{A}=A+I$ where $I$ is the identity matrix, and $\hat{D}$ is the diagonal node degree matrix of $\hat{A}$.

\section{Problem formulation}
\label{sec:problem}
In this section, we formulate the problem as a bi-level optimization. In the following, the objective function of agents and the central ISO are formulated as the first and second levels of the optimization problem, respectively. 

\subsection{Optimization model of the agents}
The aim of the generation units is to find their optimal bidding strategies which can maximize their profit while participating in the market. Let us formulate the problem as follows:

\begin{equation}
\begin{aligned}
    \max\limits_{\bidding}\SUM&\Big(\price\powera-\marginalcost\powera\Big)\\
    \text{subject to}:&\\
    & {\text{A}_{1}}:{1}\leq\bidding(t)\leq{\biddingmax},\quad\TA,\\
    \end{aligned}
    \label{eq:agent model}
\end{equation}
where the first term in the objective function $\price\powera$ is the revenue that unit $i$ obtains by participating in the market and selling its power. The second term $\marginalcost\powera$ is the marginal cost of unit $i$. Constraint $\text{A}_{1}$ limits the strategic bidding variable. If $\biddingt$ is one, it means that the agent $i$ is bidding its marginal cost and is behaving competitively. However, if it chooses a bid greater than one, it means it is participating strategically in the market.

In this market, the units send their bidding decision to the ISO. Then, ISO obtains the price of the market $\price$ and the power generation of units $\powera$. Its optimization problem is explained in the following section.

\subsection{ISO optimization model}
In this section, we derive the optimization formulation of the ISO. The goal of the ISO is to minimize the market costs while satisfying the load demand of the system. To achieve this aim, we formulate the problem as follows:

\begin{equation}
\begin{aligned}
    \min\limits_{g}\SUM\SUMI&\biddingt\marginalcost\powera\\
    \text{subject to}:&\\
    &{\text{C}_{1}}:\SUMI\powera=d(t):\quad\price,\quad\TA,\\
    &{\text{C}_{2}}:{\powermin}\leq{\powera}\leq{\powermax},\quad\N,\TA,\\
    \end{aligned}
        \label{eq:ISO}
\end{equation}
where $d(t)$ is the system demand at time $t$. Constraint $\text{C}_{1}$ expresses the demand-supply balance constraint which needs to be fulfilled at each time step. The market clearing price of the system $\lambda$ is the dual multiplier of Constraint $\text{C}_{1}$. Constraint $\text{C}_{2}$ limits the maximum and minimum power that unit $i$ can produce. 

In the following section, we explain our proposed methodology to solve problems \eqref{eq:agent model} and \eqref{eq:ISO}.

\section{Solution methodology: multi-agent RL combined with GCN}
\label{sec:solution}
In this section, we propose a solution methodology that can handle the incomplete information challenges. The proposed methodology can solve the bi-level optimization problem where agents neither know the bidding strategies of the other agents nor the price of the electrical market.

\subsection{Schematic of the methodology}
\begin{figure}[!t]
\centerline{\includegraphics[width=4in]{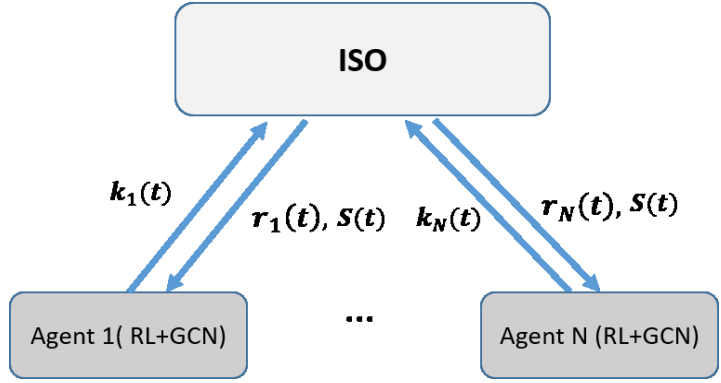}}
\caption{Schematic of the proposed solution methodology.}
\label{fig:schematic}
\end{figure}

In this problem, in the upper-level (Eq. \ref{eq:agent model}), the agents aim to maximize their revenue by obtaining their optimal bidding strategies. However, as mentioned above, the agents do not know the price of the market $\price$ which is determined by the ISO based on the bidding decisions of agents. To solve this challenge, we propose a novel deep RL learning algorithm where agents can learn the bidding strategies $\bidding$ of other agents by getting feedback from the environment. In this approach, each agent sends its bidding strategies to the ISO. ISO then clears the market and obtains the price of the market by solving Equation \ref{eq:ISO}. After that, the agents obtain their reward and based on this feedback they update their decisions. This procedure continues until they can converge to the decision that brings them the maximum profit. The schematic of this method is shown in Figure \ref{fig:schematic}.

The DRL algorithm is explained in the following section.

\subsection{Deep reinforcement learning}
As mentioned in Section \ref{sec:preliminary}, DRL applies two neural networks for estimating the value function and policy of the agents. To implement this learning algorithm, we need to define the state, action, and reward function of the agents. Let us define the state of the system as follows:

\begin{equation}
s(t)=(\lambda(t-1),d(t-1)),
\label{eq:state}
\end{equation}
which is made up of the price $\lambda$ and demand $d$ of the system at time instant $t-1$. In fact, this state includes the information of all other agents since $\lambda$ is the function of all agents' decisions. Hence, this state can describe the environment and is provided by ISO. The action of agent $i$ can be written as follows:

\begin{equation}
a_{i}(t)=\biddingt,
\end{equation}
where $1\leq\biddingt\leq\biddingmax$. To define the reward function, let us consider the objective function of agents in Equation \ref{eq:agent model}. Hence, the reward function of agent $i$ can be formulated as:

\begin{equation}
r_{i}(t)=\price\powera-\marginalcost\powera,
\label{eq:reward}
\end{equation}
which defines the revenue that the agents can obtain by selling their generated power in the market while they pay the cost of their production. In what follows, we explain the details of the algorithm and how the weights of the neural networks can be updated. 

As mentioned above, DRL estimates the value function and policy of the agents by using two neural networks, namely critic and actor networks. In the following, we drive the formulation how the weights of these neural networks are updated.

\textbf{Critic network}
The critic network is used for estimating the value function. The input of this network is the state of the system $s(t)$ and the output is the associated action-value function $Q_{i}(s(t),a_{i}(t))$. Let us define $\theta_{i}$ as the weights of the critic network which are updated in every iteration of the learning algorithm. The loss function of the neural network is defined as follows:

\begin{equation}
\begin{aligned}
L_{i}(\theta_{i})=&\frac{1}{N_{s}}\sum\limits_{j\in\mathcal{N}_{s}}\big(Q_{i}^{j}(t)-Q_{i}(s^{j}(t),a_{i}^{j}(t);\theta_{i})\big)^{2},\\
&\quad\I,
\end{aligned}
\end{equation}
where $\mathcal{N}_{s}$ is the set of samples that are used to update the weights of neural networks. The action-value function $Q_{i}$ is calculated as follows:

\begin{equation}
\begin{aligned}
Q_{i}^{j}(t)&=r_{i}^{j}(t)+\gamma\max\limits_{a_{i}^{j}(t+1)}Q_{i}(s^{j}(t+1),a_{i}^{j}(t+1),\theta_{i}'),\\
&\quad\J,
\label{eq:q_target}
\end{aligned}
\end{equation}
This equation implies that the action-value function is the summation of the current reward $r_{i}^{j}(t)$ and the discounted value of the maximum action-value function in the next time step $t+1$.

The weights of the critic network can be updated as follows:

\begin{equation}
\theta_{i}\leftarrow\theta_{i}-\eta_{i}\nabla_{\theta_{i}}L_{i}(\theta_{i}),
\label{eq:critic weight}
\end{equation}
where $\eta_{i}$ is the learning rate of the critic network. 

\textbf{Actor network}
The policy of the agents is obtained using the actor network. The state of the system is the input of the network, and the aim is to obtain the policy that can maximize the action-value function, which is formulated as:

\begin{equation}
    \pi_{i}(s(t))=\arg\max_{a_{i}(t)}{Q_{i}(s(t),a_{i}(t))}.
\end{equation}

Let us define the loss function as follows:

\begin{equation}
    L_{i}(\theta_{i}^{\mu})=-\frac{1}{N_{s}}\sum\limits_{\J}\big(Q_{i}(s^{j}(t),a_{i}^{j}(t);\theta_{i})\big)|_{a_{i}^{j}(t)=\pi_{i}(s^{j}(t),\theta_{i}^{\mu})}\big),
    \label{eq:loss actor}
\end{equation}
where $\theta_{i}^{\mu}$ is the weight of the actor network and $\pi_{i}(s^{j}(t),\theta_{i}^{\mu})$ is the policy of the agent generated by the actor network. In this formulation, by minimizing the loss function, we can obtain the policy that maximizes ${Q_{i}(s(t),a_{i}(t))}$. The weights of the network are updated as:

\begin{equation}
    \theta_{i}^\mu\leftarrow \theta_{i}^\mu-\eta_{i}^\mu\nabla_{\theta_{i}^{\mu}}L(\theta_{i}^{\mu}),
    \label{eq:actor network1}
\end{equation}
where $\eta_{i}^\mu$ is the learning rate of the actor network of unit $i$.

However, when we use deep neural networks as the critic and actor networks, we cannot integrate the network's information into the agents' decisions which is essential for the decision making. To overcome this limitation and provide the agents with additional information such as the network topology and their connection, we replace the deep neural network with the graph convolution neural network. In the next section, we explain the structure of GCN.

\begin{figure}[!t]
\centerline{\includegraphics[width=6.5in]{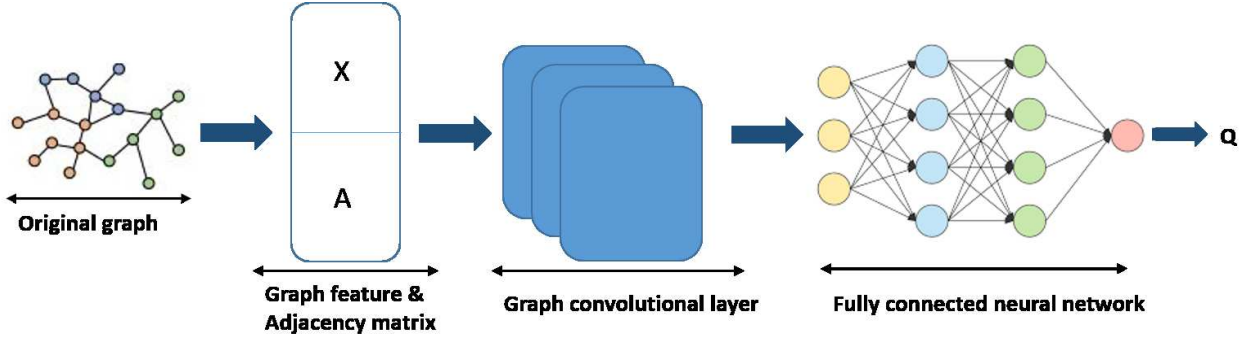}}
\caption{The schematic of the GCN critic network}
\label{fig:GCN}
\end{figure}

{\subsection{RL combined with GCN}}
In this section, we present the RL algorithm with GCN to integrate the topology of the system in the learning algorithm. In this methodology, we substitute deep feed-forward neural networks that are typically applied as actor and critic networks with GCN algorithms. The schematic of the critic network based on GCN is shown in Figure \ref{fig:GCN}. As we can see, the architecture of this network can be divided into two parts: 1) feature extraction and 2) policy approximation. The input of the GCN is the state of the system which consists of the features of  node $X$ and the adjacency matrix $A$ which can be described as the tuple of ${\{X, A}\}$. The state of the system Equation \ref{eq:state} can be considered as the features of the nodes.

The policy can be approximated by GCN in two parts. The first part extracts the features and the second part approximates the policy.  

The iteration algorithm of RL with GCN is outlined in Algorithm \ref{algorithm2}.

\begin{algorithm}
	\caption{RL algorithm with GCN for bidding strategy} 
	\label{algorithm2}
	\begin{algorithmic}[1] 
		\STATE {\textbf{Input}}: $\gamma,\biddingmax,\powermin,\powermax\theta_{i}\eta_{i}$,$\theta^{\mu}_{i},\eta^{\mu}_{i}, \N$\\ 
        \FOR{$episode = 1,\ldots,\mathcal{M}$}
                \STATE Initialize the system state $s(t)=(d(t), \lambda(t))$
        \STATE Initialize the system input ${s(t), A}$
        \STATE $\mathcal{T}={\{1+(episode-1)T,\cdots,(episode)T}\}$
        \FOR{$t \in\mathcal{T}$}
        \STATE Obtain the action from the actor network $a_{i}(t)=\max(\min(\mu_{i}(s(t),\theta_{i}^{\mu}),\biddingmax),1)$,\quad $\N$
        \STATE Obtain the market clearing price $\lambda(t)$ and units' power generation $g_{i}(t)$, $\N$, using \eqref{eq:ISO}
        \STATE Observe the next state $s(t+1)=(d(t+1), \lambda(t+1))$ 
        \STATE Calculate the reward function $r_{i}(t)$, $\N$, using \eqref{eq:reward}
        \STATE Store the transition $(s(t),s(t+1),a_{i}(t),r_{i}(t))$, $\N$
        \STATE Randomly sample $(s^{j}(t),s^{j}(t+1),a^{j}_{i}(t),r^{j}_{i}(t))$, $\N$, $\J$
        \STATE Obtain  $Q_{i}^{j}(t)$, $\N$, $\J$, using \eqref{eq:q_target}
        \STATE Update the  GCN critic networks' weights $\theta_{i}$, $\N$, using \eqref{eq:critic weight}
        \STATE Update the GCN actor networks' weights $\theta_{i}^{\mu}$, $\N$, using \eqref{eq:actor network1}
        \ENDFOR
                \ENDFOR
	\end{algorithmic}
\end{algorithm}

\section{Case study evaluations}
\label{sec:case study}
In this section, we evaluate the proposed algorithm on the IEEE 30-bus system. We consider different scenarios and compare the performances of RL combined with GCN with the RL algorithm. We also examine the generalization ability of the proposed approach by testing the trained model on IEEE 39-bus system.

\subsection{Overview of the system model}
IEEE 30-bus system consists of $6$ generation units (agents) and the schematic of this system is shown in Figure \ref{fig:schematic}. In this system, we assume that the maximum bidding $\biddingmax$ is $2$ (Constraint $A_1$ in \ref{eq:agent model}) and means that agent participates in the market strategically. The agents' parameters are shown in Table \ref{tab:generation}. The same parameters of the network as in \cite{bompard2006network} were used.

\begin{figure}[!t]
\centerline{\includegraphics[width=4in]{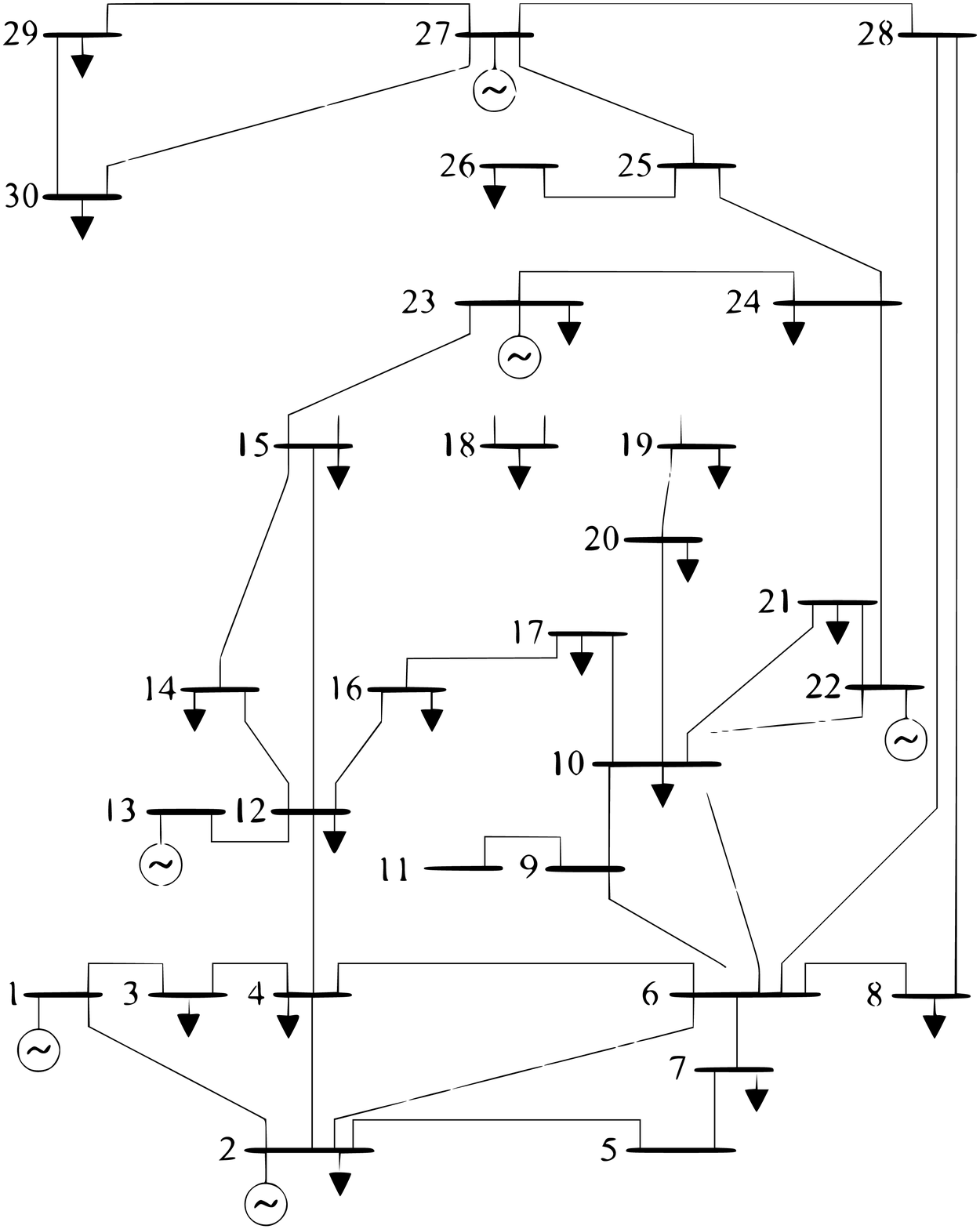}}
\caption{Schematic of IEEE 30-bus system with $6$ generation units.}
\label{fig:profit1}
\end{figure}

\begin{table}
  \caption{Generation units' parameters of IEEE 30-bus system}
  \centering
 \begin{tabular}{c | c c c c} 
Generation unit&  $\marginalcost$  & $\powermax$ &   $\powermin$ &$c_{i}$ \\
& $[\frac{\$}{MWh}]$  & $[MW]$  & $[MW]$ & $[{\$}]$ \\
 \hline
 1 &  2& 80 &5 & 120 \\ 
 \hline
 2 & 1.75 & 80 &5 &135  \\
 \hline
 3 & 1& 50 &5 &142 \\
 \hline
 4 &  3.25& 50& 5 &125\\
 \hline
 5 & 3 &35 & 5 &175\\
  \hline
 6 & 3 &40 & 5 &165 \\
\end{tabular}
  \label{tab:generation}
\end{table}
To implement the proposed Algorithm \ref{algorithm2}, we consider $50$ episodes in which the agents can converge to their decisions and each episode consist of $30$ days. In addition, we assume that $\gamma$, $\eta_{i}$, and $\eta^{\mu}_{i}$ of Algorithm \ref{algorithm2}, are $0.9$, $0.1$, and $0.1$, respectively. For the GCN network, we use two hidden layers with the Rectified Linear Unit (ReLU) activation function. Each hidden layer of GCN has $16$ ReLU. In addition, for the feed-forward part of the GCN, we use two hidden layers with $15$ and $10$ ReLU as well. The initial value of the weights of neural networks is initialized randomly, in the range between $1$ and $2$. In the following, we evaluate the performance of the proposed algorithm on the one hand to evaluate its ability to learn optimal policies in different scenarios and on the other hand its ability to generalize to new architectures.  We first implement the algorithm on the IEEE 30-bus system with different generation capacities and investigate the performance of the algorithm and its sensitivity on the capacity of the agents. In the second step, we evaluate the generalization ability of the RL combine with GCN with DRL. For this, we train the algorithm on the IEEE 30-bus system case study and then apply the trained algorithm to the IEEE 39-bus system case study. The 39 bus system has $9$ generation units and the agents' parameters are shown in Table \ref{tab:generation2}. In addition, we evaluate the generalization capability of the algorithm when the system is affected by a fault and some of the buses are disconnected. In the following, the results of all these scenarios are presented.

\begin{table}
  \caption{Generation units' parameters of IEEE 39-bus system}
  \centering
 \begin{tabular}{c | c c c c} 
Generation unit&  $\marginalcost$  & $\powermax$ &   $\powermin$ &$c_{i}$ \\
& $[\frac{\$}{MWh}]$  & $[MW]$  & $[MW]$ & $[{\$}]$ \\
 \hline
 1 &  2& 80 &5 & 120 \\ 
 \hline
 2 & 1.75 & 80 &5 &135  \\
 \hline
 3 & 1& 50 &5 &142 \\
 \hline
 4 &  3.25& 50& 5 &125\\
 \hline
 5 & 3 &35 & 5 &175\\
  \hline
 6 & 3 &40 & 5 &165 \\
\end{tabular}
  \label{tab:generation2}
\end{table}

\subsection{Varying the generation capacity}
In this section, we compare the average profit of units for three different scenarios. These two scenarios are expressed as Table \ref{tab:table2}. 

\begin{table}
  \caption{Varying the generation capacity}
  \centering
 \begin{tabular}{c | c c c} 
Generation unit&  First scenario  & Second scenario & Third scenario \\
 &  $[MW]$  & $[MW]$  & $[MW]$ \\
 \hline
 1 &  80& 150 & 80\\ 
 \hline
 2 & 80 & 150 & 80\\
 \hline
 3 & 50& 50 & 50\\
 \hline
 4 &  50& 50& 50\\
 \hline
 5 & 35 &30 & 35\\
  \hline
 6 & 40 &40  & 100\\
\end{tabular}
  \label{tab:table2}
\end{table}

\begin{figure}[!t]
\centerline{\includegraphics[width=4.5in]{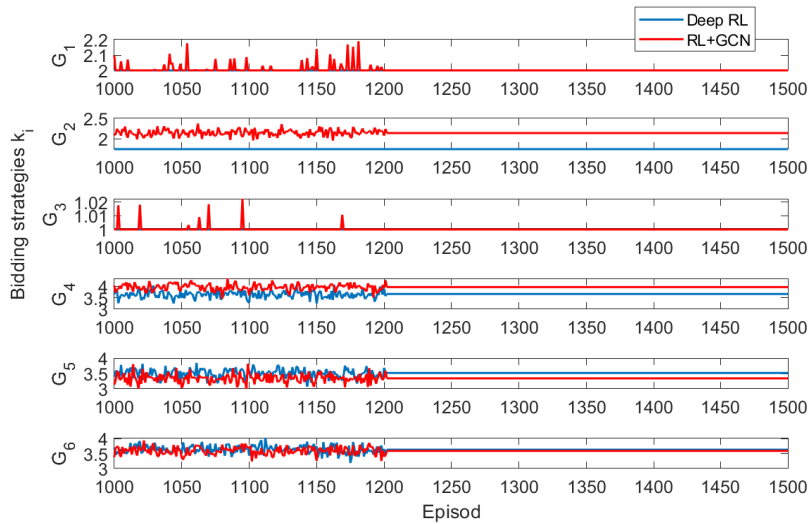}}
\caption{Comparison the bidding strategies of DRL with RL+GCN for IEEE 30-bus systems with the generation capacity of $[80,80,50,50,35,40]$.}
\label{fig:bidding1}
\end{figure}

In the first scenario, the capacity of units is the same as Table \ref{tab:generation}. The average profit and bidding strategies of the units are shown in Figures \ref{fig:profit1} and \ref{fig:bidding1}, respectively.

\begin{figure}[!t]
\centerline{\includegraphics[width=4in]{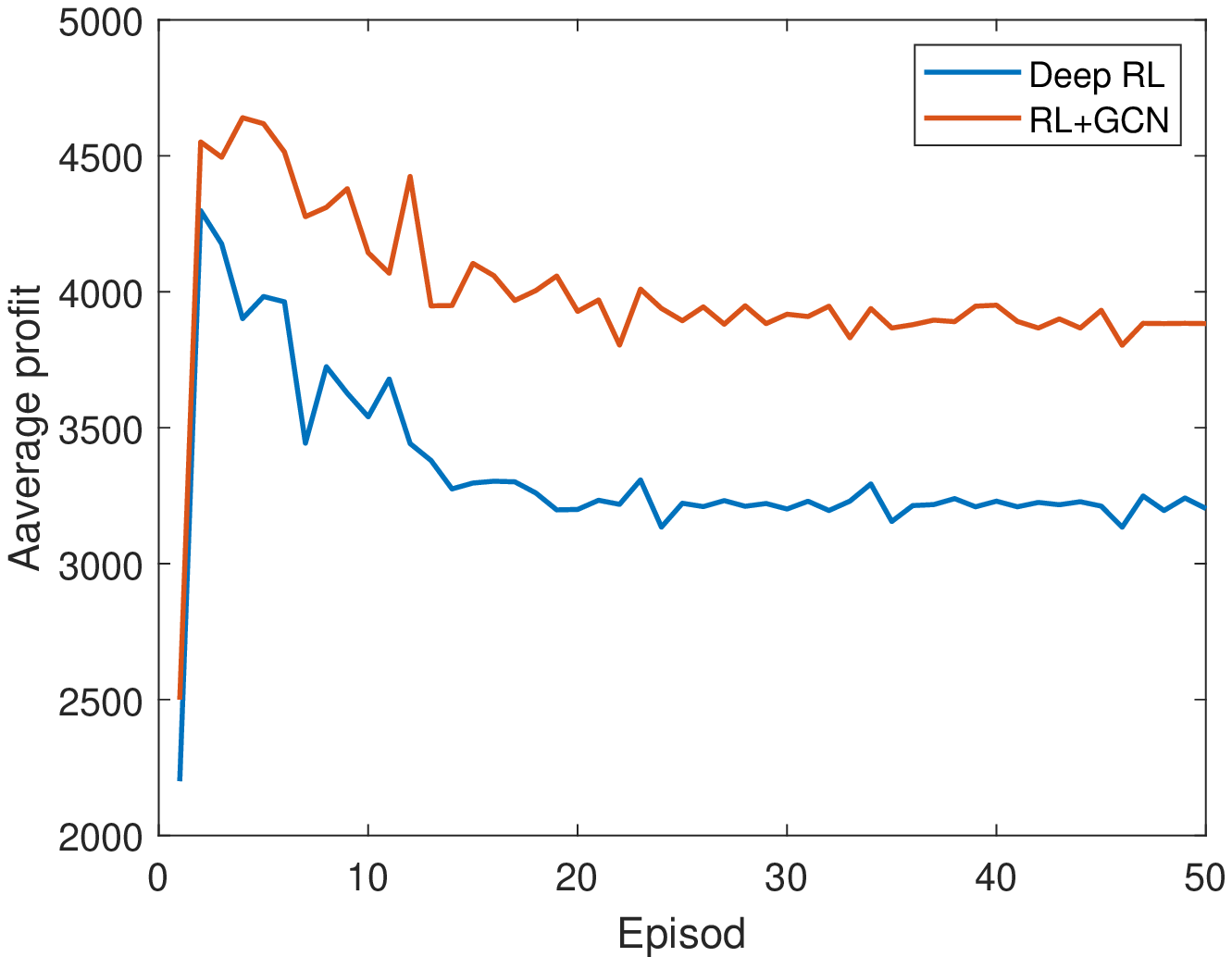}}
\caption{Comparison the average profit of DRL with RL+GCN for IEEE 30-bus systems with the generation capacity of $[80,80,50,50,35,40]$.}
\label{fig:profit1}
\end{figure}

\begin{figure}[!t]
\centerline{\includegraphics[width=4in]{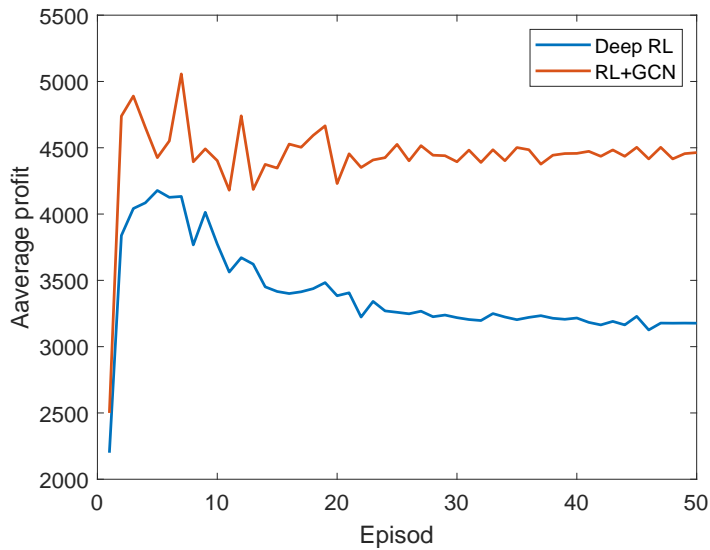}}
\caption{Comparison the average profit of DRL with RL+GCN for IEEE 30-bus systems with the generation capacity of $[150,150,50,50,30,40]$. }
\label{fig:profit2}
\end{figure}

\begin{figure}[!t]
\centerline{\includegraphics[width=4.5in]{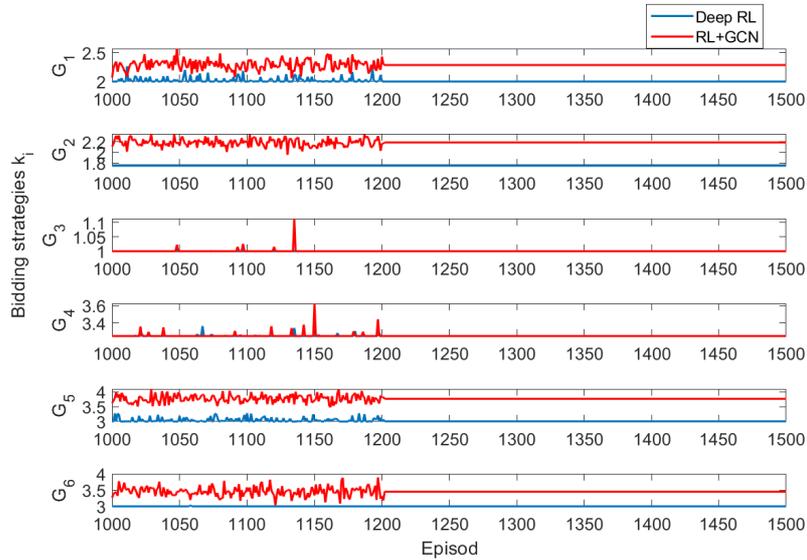}}
\caption{Comparison of the bidding strategies of DRL with RL+GCN for IEEE 30-bus systems with the generation capacity of $[150,150,50,50,30,40]$.}
\label{fig:bidding2}
\end{figure}

In the second scenario, we change the capacity of generation units. In this scenario, we assume that the first and second units have the higher capacity compare to the first scenario. We have chosen this case to see how GCN can help the units which have the highest impact on the price of the market. The average profit and bidding strategies of units are shown in Figures \ref{fig:profit2} and \ref{fig:bidding2}, respectively.

\begin{figure}[!t]
\centerline{\includegraphics[width=4in]{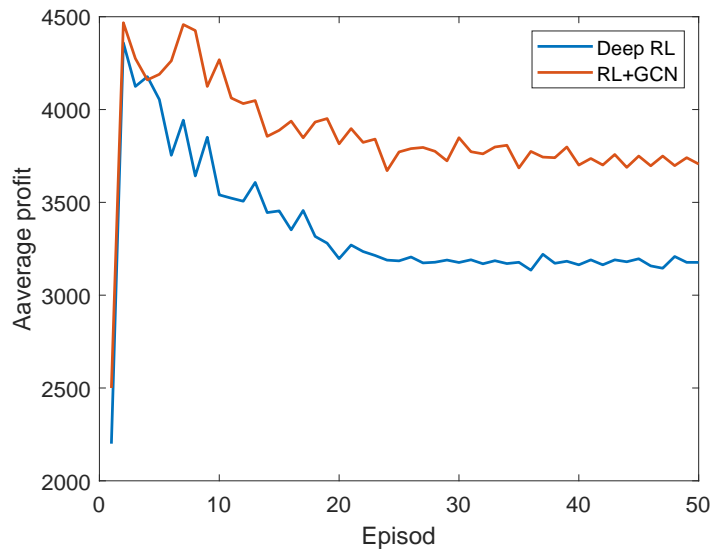}}
\caption{Comparison the average profit of DRL with RL+GCN for IEEE 30-bus systems with the generation capacity of $[80,80,50,50,35,40]$. }
\label{fig:profit3}
\end{figure}

\begin{figure}[!t]
\centerline{\includegraphics[width=6.6in]{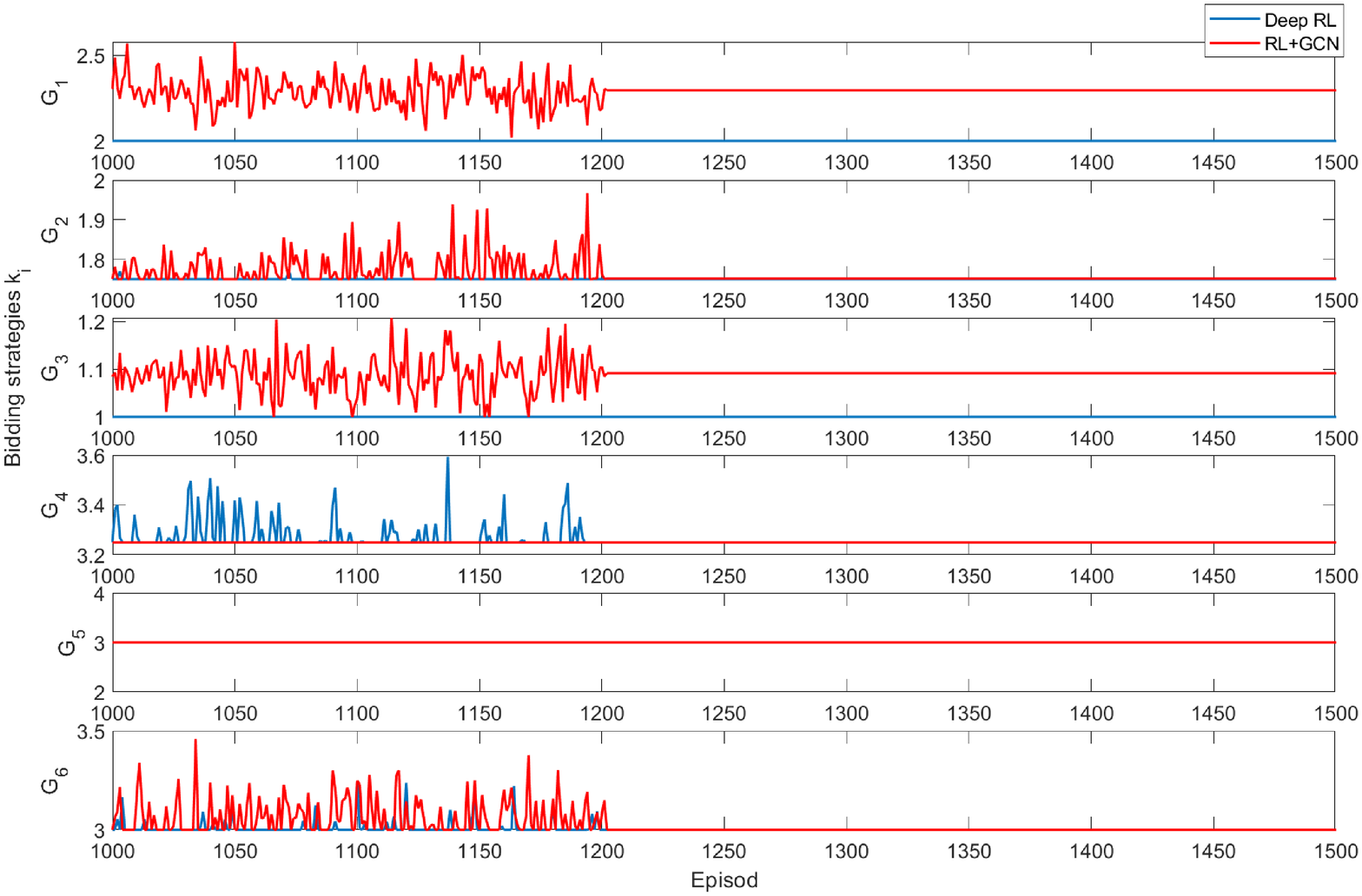}}
\caption{Comparison of the bidding strategies of DRL with RL+GCN for IEEE 30-bus systems with the generation capacity of $[80,80,50,50,35,40]$.}
\label{fig:bidding3}
\end{figure}

The average profit and bidding strategies of units for the third scenario in which the capacity of unit $6$ increases, are shown in Figures \ref{fig:profit3} and \ref{fig:bidding3}, respectively.

As we can see from Figures \ref{fig:profit1}, the average profit of the RL with GCN for fis about $3800$. The average profit of the RL with GCN for the second scenario is around $4500$. Hence, we can conclude that when the generating capacity of Agents 1 and 2 increases, the average profit of RL with GCN also increases. However, the average profit of DRL for the first and second scenarios are about $3300$ and cannot significantly increase. These results arise because GCN can take into account the topology of the system and thus help agents $1$ and $2$ to increase their bids form $2$ to $2.3$ and $1.8$ to $2.3$, respectively (as Figure \ref{fig:bidding2}), resulting in an increase in the market price. In other words, due to the topology of the network, these two agents can win the market and even generate electricity by increasing their bids. So, as the price of the market increases, the reward of all units increases, and the average profit also increases.

\subsection{Generalization capability: different network typologies}
\begin{figure}[!t]
\centerline{\includegraphics[width=4in]{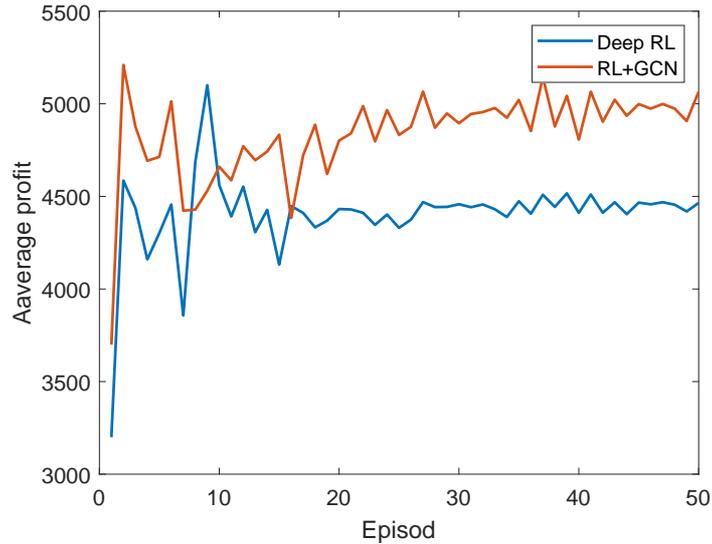}}
\caption{Comparison the average profit over all units of DRL with RL+GCN for IEEE 39-bus systems.}
\label{fig:compare}
\end{figure}

In this section, we evaluate the generalization ability of RL with GCN and DRL. To achieve this goal, we first train these two methods for the IEEE 30-bus system with the parameters of Table \ref{tab:generation} and then test the learned model on the IEEE 39-bus systems. The average rewards of these two approaches are shown in figure \ref{fig:compare}. According to the results visualised in this figure, RL in combination with GCN algorithm can achieve a higher profit compared to DRL. From this, we can conclude that the generalization ability of this method is more than DRL. The reason for this result is that GCN can integrate the topology of the system in the learning algorithm, so it can better adapt to the new topology than DRL.

\subsection{Generalization capability: bus disconnection}
In this part, we examine the generalization capability of RL with GCN for IEEE 30-bus system with the parameters of Table \ref{tab:generation} when a fault happens and some of the lines are disconnected. We assume three scenarios in which 3, 5, and 10 lines of Figure \ref{fig:schematic} are disconnected. In the first scenario the lines between nodes 3-4, 8-6, and 10-21 are disconnected. For the second case, we assume that there are the disconnection between nodes 3-4, 7-6, 16-17, 10-21, and 24-25. In the last case, we assume that there are not any connections between nodes 1-2, 3-4, 4,6- 2-5, 1-3, 2-4, 4-12, 29-30, 27-28, and 19-20. The average profit of agents for all the scenarios are shown in Table \ref{tab:table3}.

\begin{table}
  \caption{Average profit of the Rl with GCN and DRL algorithms for IEEE 30-bus system with disconnected bus}
  \centering
 \begin{tabular}{c | c c} 
Number of disconnected bus&  RL+GCN & DRL\\
 \hline
 3 & 3900 & 3650\\ 
 \hline
 5 & 4040 & 3300 \\
 \hline
 10 &3800  &3200 \\
\end{tabular}
  \label{tab:table3}
\end{table}
As we can see from Table \ref{tab:table3}, the average reward of RL with GCN is higher than DRL for all the cases. Since, GCN can consider the topology of the system, it can help agents to choose the strategies that will result in a higher profit. In addition, when the number of disconnected bus is increasing from $3$ to $10$ the average profit of DRL is decreasing significantly from $3650$ to $3200$, however, this is not the case for RL with GCN and the profit does not change significantly. These results demonstrate that the GCN can adapt to the new topology better than DRL.

\section{Conclusions}
\label{sec:conclusion}
In this paper, we have proposed a framework combining RL with GCN for solving the problem of bidding strategies of generation units in the electrical market. This approach can overcome the challenges of incomplete information and system uncertainties. In addition, the proposed algorithm can help the agents learn the topology of the system and provide more information to the agents to make their decisions. The results show that RL with GCN has a better performance compared to DRL that does not take the topology into consideration. In addition, it has a better generalization ability when the topology of the system changes or some buses are disconnected.

In future work, it would be interesting to integrate more uncertainties to the system such as considering the wind turbine system. Also, some other constraints of the network and the agents can be taken into account in modeling the system. Furthermore, it would also be interesting to evaluate the proposed framework on larger networks  implement the proposed algorithm in a larger system with more nodes and examine the performance of the algorithm.

\section*{Acknowledgement}
\vspace{-3pt}
The contribution of Pegah Rokhforoz was funded by the Hasler foundation agency.
\vspace{-3pt}

\bibliography{bib_items}
\bibliographystyle{ieeetr}
\end{document}

%% file: Style_P.tex
\makeatletter
\def\@settitle{\begin{center}%
		\baselineskip14\p@\relax
		\normalfont\LARGE\scshape\bfseries
		\@title
	\end{center}%
}
\makeatother

\makeatletter

\def\subsection{\@startsection{subsection}{2}%
	\z@{.5\linespacing\@plus.7\linespacing}{.5\linespacing}%
	{\normalfont\bfseries}}

\def\subsubsection{\@startsection{subsubsection}{3}%
	\z@{.5\linespacing\@plus.7\linespacing}{.5\linespacing}%
	{\normalfont\itshape}}

\usepackage[usenames, dvipsnames]{color}
\definecolor{darkblue}{rgb}{0.0, 0.0, 0.45}

\usepackage[colorlinks	= true,
			raiselinks	= true,
			linkcolor	= darkblue, 
			citecolor	= Mahogany,
			urlcolor	= ForestGreen,
			pdfauthor	= {},
			pdftitle	= {},
			pdfkeywords	= {},
			pdfsubject	= {},
			plainpages	= false]{hyperref}

\usepackage{dsfont,amssymb,amsmath,graphicx} 
\usepackage{amsfonts,dsfont,mathtools,mathrsfs,amsthm} 
\mathtoolsset{showonlyrefs=true}
\usepackage[amssymb, thickqspace]{SIunits}
\usepackage{fancyhdr,setspace}

\usepackage[numbers,compress]{natbib}
\usepackage{nicefrac}       
\usepackage{microtype}      
\usepackage{tabto}

\usepackage{enumerate,enumitem}
 \usepackage{caption}

\usepackage{bbm}

\allowdisplaybreaks
\date{\today}

%% file: command.tex
\theoremstyle{theorem}

\theoremstyle{remark}

\newcommand{\powermax}{g_{i}^{\max}}
\newcommand{\powermin}{g_{i}^{\min}}

\newcommand{\powera}{g_{i}(t)}

\newcommand{\marginalcost}{\lambda_{i,m}}

\newcommand{\price}{\lambda(t)}

\newcommand{\bidding}{k_{i}}
\newcommand{\biddingt}{k_{i}(t)}

\newcommand{\biddingmax}{k_{i}^{\max}}

\newcommand{\SUM}{\sum\limits_{t\in\mathcal{T}}}
\newcommand{\TA}{\forall{t\in\mathcal{T}}}

\newcommand{\SUMI}{\sum\limits_{i\in\mathcal{N}}}

\newcommand{\J}{\forall j\in\mathcal{N}_{s}}
\newcommand{\I}{\forall i\in\mathcal{N}}

\newcommand{\N}{{\forall i\in\mathcal{N}}}